\documentclass{article}
\usepackage{ijcai19}
\usepackage{float}
\usepackage{times}
\usepackage{soul}
\usepackage{hyperref}
\usepackage[utf8]{inputenc}
\usepackage[small]{caption}
\usepackage{graphicx}
\usepackage{amsmath}
\usepackage{booktabs}
\usepackage{booktabs}

\usepackage[super]{nth}
\usepackage{graphicx}
\title{SocialNLP EmotionX 2019 Challenge Overview: \\
Predicting Emotions in Spoken Dialogues and Chats}

\author{
Boaz Shmueli$^1,^2$\footnote{Contact Author}\and
Lun-Wei Ku$^3$\\
\affiliations
$^1$Social Networks and Human-Centered Computing, Taiwan International Graduate Program (TIGP)\\
Institute of Information Science, Academia Sinica\\
$^2$National Tsing Hua University\\
$^3$Institute of Information Science, Academia Sinica\\
\emails
\{shmueli, lwku\}@iis.sinica.edu.tw
}

\begin{document}

\maketitle

\begin{abstract}
We present an overview of the EmotionX 2019 Challenge, held at the 7th International Workshop on 
Natural Language Processing for Social Media (SocialNLP), in conjunction with IJCAI 2019. 
The challenge entailed predicting emotions in spoken and chat-based dialogues using augmented EmotionLines datasets. 
EmotionLines contains two distinct datasets: the first includes excerpts from a US-based TV sitcom episode scripts (Friends) and the second contains real online chats (EmotionPush). 
A total of thirty-six teams registered to participate in the challenge. 
Eleven of the teams successfully submitted their predictions for performance evaluation. 
The top-scoring team achieved a micro-F1 score of 81.5\% for the spoken-based dialogues (Friends) and 79.5\% for the chat-based dialogues (EmotionPush). 
\end{abstract}

\section{Introduction}
Emotions are a central component of our existence as human beings, and are manifested by physiological and psychological changes that often affect behavior and action. Emotions involve a complicated interplay of mind, body, language, and culture \cite{Bazzanella2004}.

Detecting and recognizing emotions is a difficult task for machines. 
Nevertheless, following the successful use of computational linguistics to analyze sentiment in texts, there is growing interest in the more difficult task of the automatic detection and classification of emotions in texts.

The detection of emotions in text is a complicated challenge for multiple reasons: first, emotions are complex entities, and no universally-agreed upon psychological model of emotions exists. 
Second, isolated texts convey less information compared to a complete human interaction in which emotions can be detected from the other person's facial expressions, listening to their tone of voice, etc. However, due to important applications in fields such as psychology, marketing, and political science, research in this topic is now expanding rapidly \cite{seyeditabari2018emotion}.

In particular, dialogue systems such as those available on social media or instant messaging services are rich sources of textual data and have become the focus of much attention. Emotions of utterances within dialogues can be detected more precisely due to the presence of more context.
For example, a single utterance (``OK!'') might convey different emotions (happiness, anger, surprise), depending on its context. Taking all this into consideration, in 2018 the EmotionX Challenge asked participants to detect emotions in complete dialogues \cite{hsu2018socialnlp}. Participants were challenged to classify utterances using Ekman's  well-known theory of six basic emotions (sadness, happiness, anger, fear, disgust, and surprise) \cite{ekman1999basic}.

For the 2019 challenge, we built and expanded upon the 2018 challenge. We provided an additional 20\% of data for training, as well as augmenting the dataset using two-way translation. The metric used was micro-F1 score, and we also report the macro-F1 score.

A total of thirty-six teams registered to participate in the challenge. 
Eleven of the teams successfully submitted their data for performance evaluation, and seven of them submitted technical papers for the workshop.
Approaches used by the teams included deep neural networks and SVM classifiers. 
In the following sections we expand on the challenge and the data. We then briefly describe the various approaches used by the teams, and conclude with a summary and some notes. Detailed descriptions of the various submissions are available in the teams' technical reports.

\section{Datasets}
The two datasets used for the challenge are Friends and EmotionPush, part of the EmotionLines corpus \cite{chen2018emotionlines}. 
The datasets contain English-language dialogues of varying lengths. 
For the competition, we provided 1,000 labeled dialogues from each dataset for training, and 240 unlabeled dialogues from each dataset for evaluation. 
The Friends dialogues are scripts taken from the American TV sitcom (1994-2004). 
The EmotionPush dialogues are from Facebook Messenger chats by real users which have  been anonymized to ensure user privacy\footnote{To guarantee high level of anonymization, an automatic Named Entity Recognition (NER) replacement was followed by careful manual inspection and correction}. For both datasets, dialogue lengths range from 5 to 24 lines each. A breakdown of the lengths of the dialogues is shown in Table \ref{tab:buckets}.  

We employed workers using Amazon Mechanical Turk (aka AMT or MTurk) to annotate the dialogues \cite{callison2010creating}.
Each complete dialogue was offered as a single MTurk Human Intelligence Task (HIT), within which each utterance was read and annotated for emotions by the worker. Each HIT was assigned to five workers. To ensure  workers were qualified for the annotation task, we set up a number of  requirements: workers had to be from an English-speaking country (Australia, Canada, Great Britain, Ireland, New Zealand, or the US), have a high HIT approval rate (at least 98\%), and have already performed a minimum of 2,000 HITs.

% Please add the following required packages to your document preamble:
% \usepackage{booktabs}
% Please add the following required packages to your document preamble:
% \usepackage{booktabs}
\begin{table}[hpt!]
\centering
\begin{tabular}{@{}ccc@{}}
\toprule
 \textbf{Friends} & \textbf{EmotionPush} & \textbf{Both}\\ 
\toprule
  0.326 & 0.342 & 0.345\\ 

\bottomrule
\end{tabular}
\caption{Reliability of Agreement ($\kappa$) }
\label{tab:irr}
\end{table}
In the datasets, each utterance is accompanied by an \textit{annotation} and \textit{emotion}.
The annotation contains the raw count of votes for each emotion by the five annotators, with the order of the emotions being Neutral, Joy, Sadness, Fear, Anger, Surprise, Disgust. 
For example, an annotation of ``2000030'' denotes that two annotators voted for ``neutral'', and three voted for ``surprise''. 

The labeled emotion is calculated using the absolute majority of votes. 
Thus, if a specific emotion received three or more votes, then that utterance is labeled with that emotion. If there is no majority vote, the utterance is labeled with ``non-neutral'' label. In addition to the utterance, annotation, and label, each line in each dialogue includes the speaker's name (in the case of EmotionPush, a speaker ID was used).
The emotion distribution for Friends and EmotionPush, for both training and evaluation data, is shown in Table \ref{tab:utterances}.
% Please add the following required packages to your document preamble:
% \usepackage{booktabs}
\begin{table*}[h]
\centering
\begin{tabular}{@{}ll|r|rrrrrrrr@{}}
\multicolumn{1}{c}{\textbf{}} & \multicolumn{1}{c|}{\textbf{}} & \multicolumn{1}{c|}{\textbf{\# utterances}} & \multicolumn{1}{c}{\textbf{non-neutral}} & \multicolumn{1}{c}{\textbf{neutral}} & \multicolumn{1}{c}{\textbf{joy}} & \multicolumn{1}{c}{\textbf{sadness}} & \multicolumn{1}{c}{\textbf{anger}} & \multicolumn{1}{c}{\textbf{disgust}} & \multicolumn{1}{c}{\textbf{fear}} & \multicolumn{1}{c}{\textbf{surprise}} \\ \midrule
\textbf{EmotionPush}          & \textbf{Train}                 & 14742                                       & 1418                                     & 9855                                 & 2100                             & 514                                  & 140                                & 106                                  & 42                                & 567                                   \\
\textbf{}                     & \textbf{Test}                  & 3536                                        & 394                                      & 2146                                 & 601                              & 110                                  & 27                                 & 51                                   & 18                                & 189                                   \\
\textbf{}                     & \textbf{Subtotal}              & 18278                                       & 1812                                     & 12001                                & 2701                             & 624                                  & 167                                & 157                                  & 60                                & 756                                   \\ \midrule
\textbf{Friends}              & \textbf{Train}                 & 14503                                       & 2772                                     & 6530                                 & 1710                             & 498                                  & 759                                & 331                                  & 246                               & 1657                                  \\
\textbf{}                     & \textbf{Test}                  & 3296                                        & 952                                      & 1035                                 & 505                              & 121                                  & 141                                & 58                                   & 37                                & 447                                   \\
\textbf{}                     & \textbf{Subtotal}              & 17799                                       & 3724                                     & 7565                                 & 2215                             & 619                                  & 900                                & 389                                  & 283                               & 2104                                  \\ \midrule
\textbf{Both}                 & \textbf{Train}                 & 29245                                       & 4190                                     & 16385                                & 3810                             & 1012                                 & 899                                & 437                                  & 288                               & 2224                                  \\
\textbf{}                     & \textbf{Test}                  & 6832                                        & 1346                                     & 3181                                 & 1106                             & 231                                  & 168                                & 109                                  & 55                                & 636                                   \\ \midrule
\textbf{}                     & \textbf{Total}                 & \textbf{36077}                              & \textbf{5536}                            & \textbf{19566}                       & \textbf{4916}                    & \textbf{1243}                        & \textbf{1067}                      & \textbf{546}                         & \textbf{343}                      & \textbf{2860}       
\end{tabular}
\caption{Emotion Label Distribution}
\label{tab:utterances}
\end{table*}

We used Fleiss' kappa measure to assess the reliability of agreement between the annotators \cite{fleiss1971measuring}. The value for $\kappa$-statistic is  $0.326$ and $0.342$ for Friends and EmotionPush, respectively. For the combined datasets the value of the $\kappa$-statistic is $0.345$.\footnote{We used the \textit{irr} package for R.}

Sample excerpts from the two datasets, with their annotations and labels, are given in Table \ref{tab:dialogues}.
% Please add the following required packages to your document preamble:
% \usepackage{booktabs}
\begin{table}[t]
\begin{tabular}{@{}cccccc@{}}
\toprule
\multicolumn{2}{c|}{} & \multicolumn{2}{c}{\textbf{Friends}} & \multicolumn{2}{c}{\textbf{EmotionPush}} \\ \midrule
Bucket  & \multicolumn{1}{c|}{Lines} & Train         & Test         & Train           & Test          \\ \midrule
1       & \multicolumn{1}{c|}{5-9}   & 250           & 50           & 250             & 60            \\
2       & \multicolumn{1}{c|}{10-14} & 250           & 90           & 250             & 60            \\
3       & \multicolumn{1}{c|}{15-19} & 250           & 70           & 250             & 60            \\
4       & \multicolumn{1}{c|}{20-24} & 250           & 30           & 250             & 60            \\ \midrule
\multicolumn{2}{r|}{\textit{Dialogues}}  & 1000          & 240          & 1000            & 240  \\ \midrule
\multicolumn{2}{r|}{\textit{Utterances}}  & 14503          & 3296          & 14742           & 3536 \\ 
\bottomrule       
\end{tabular}
\caption{Dialogue Length Distribution and Number of Utterances}
\label{tab:buckets}
\end{table}

\paragraph{Augmentation}  
NLP tasks require plenty of data. Due to the relatively small number of samples in our datasets, we added more labeled data using a technique developed in \cite{ostyakov2018} that was used by the winning team in Kaggle's Toxic Comment Classification Challenge \cite{lee2018}. 
The augmented datasets are similar to the original data files, but include additional machine-computed utterances for each original utterance. 
We created the additional utterances using the Google Translate API. Each original utterance was first translated from English into three target languages (German, French, and Italian), and then translated back into English. The resulting utterances were included together in the same object with the original utterance. These ``duplex translations'' can sometimes result in the original sentence, but many times variations are generated that convey the same emotions.

\begin{table}[htbp]
\centering
\resizebox{\columnwidth}{!}{%

\begin{tabular}{ll} 
\toprule
\textbf{Source} & \textbf{Utterance}                            \\ 
\midrule
Original        & "Wow! That's ah, that's pretty nice!"         \\
English $\rightleftharpoons$ German     & "Impressive! That's ah, that's pretty nice!" \\
English $\rightleftharpoons$  French      & "Wow! It's ah, it's pretty cool!"             \\
English $\rightleftharpoons$  Italian     & "Wow! It's ah, it's cute!" \\ 
\bottomrule
\end{tabular}
}
\caption{Example of Augmented Utterance}
\label{tab:augmented}
\end{table}

Table \ref{tab:augmented} shows an example utterance (labeled with ``Joy'') after augmentation. 
% Please add the following required packages to your document preamble:
% \usepackage{booktabs}
\begin{table*}[t]
\centering
%\resizebox{\linewidth}{!}
{%
\begin{tabular}{@{}llll@{}}
\toprule
\textbf{Speaker} & \textbf{Utterance} & \textbf{Annotation} & \textbf{Emotion} \\ 
\midrule
Phoebe           & ``Ohh, that's too bad.''                                     &20\textbf{3}0000& sadness          \\
Ross             & ``No, I-I'm saying I liked her.''                            & 2010110         & non-neutral      \\
Phoebe           & ``Yeah, y'know what, there are other fish in the sea.''          &\textbf{4}010000      & neutral          \\
Ross             & ``Pheebs, I think she's great. Okay? We're going out again.''     &2100110     & non-neutral      \\
Phoebe           & ``Okay, I hear you! Are you capable of talking about any thing else?'' &000020\textbf{3}& disgust \\ \midrule
1167038771          & ``why is that poster so bad''   & 200000\textbf{3} & disgust             \\
1220662692          & ``what do you mean its great''  & 10000\textbf{4}0 & surprise     \\
1167038771          & ``oh''             & \textbf{3}000020 & neutral              \\
1167038771          & ``Can You Really Trust Anyone'' & 2010011 & non-neutral  \\
1220662692          & ``lol''            & 0\textbf{5}00000 & joy                 \\
1167038771          & ``Can You?''       & \textbf{3}001010 & neutral                 \\  
1220662692          & ``i guess not''    & 20\textbf{3}0000 & sadness               \\
\toprule
\end{tabular}
}
\caption{Dialogue Excerpts from Friends (top) and EmotionPush (bottom)}
\label{tab:dialogues}
\end{table*}
\section{Challenge Details}
A dedicated website for the competition was set up\footnote{https://sites.google.com/view/emotionx2019}. 
The website included instructions, the registration form, schedule, and other relevant details. 
Following registration, participants were able to download the training datasets.

\begin{figure*}[h]
\centering
\includegraphics[width=\linewidth]{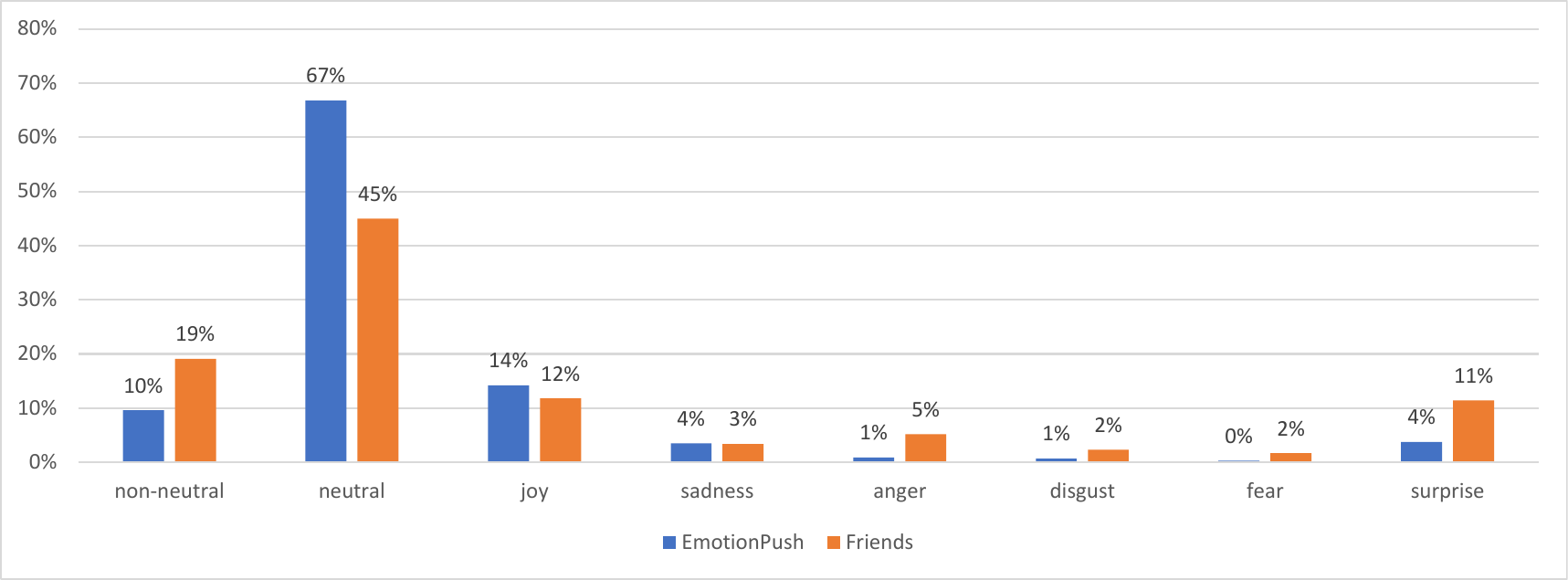}
\caption{Emotion Label Distribution (Training Datasets)}
\label{figure:labels}
\end{figure*}

The label distribution of emotions in our data are highly unbalanced, as can be seen in Figure \ref{figure:labels}. 
Due to the small number of three of the labels, participants were instructed to use only four emotions for labels: joy, sadness, anger, and neutral. Evaluation of submissions was done using only utterances with these four labels. Utterances with labels other than the above four (i.e., surprise, disgust, fear or non-neutral) were discarded and not used in the evaluation.
Scripts for verifying and evaluating the submissions were made available online\footnote{http://bit.ly/emotionx2019-verify}. 
We used micro-F1 as the comparison metric.

\section{Submissions}
% Please add the following required packages to your document preamble:
% \usepackage{booktabs}
\begin{table*}[t]
\centering
\begin{tabular}{@{}lllcrrrr@{}}
\toprule
\textbf{Rank} & \textbf{Team} & \textbf{Features} & \textbf{F1-micro} & \textbf{Neutral} & \textbf{Joy} & \textbf{Sadness} & \textbf{Anger} \\ 
\midrule
1 & IDEA & BERT, speaker, transfer, context & 81.5 & 87.3 & 75.5 & 59.6 & 69.8 \\
2 & KU & BERT, ensemble, context&  79.5 & 86.0 & 72.0 & 51.4 & 65.6 \\
3 & HSU & BERT        & 79.1 & 85.4 & 73.6 & 55.6 & 65.0 \\
4 & AlexU & BERT, majority classifier & 77.0 & 84.5 & 72.3 & 58.0 & 59.7 \\
5 & Antenna     & BERT+GRU   & 75.2 & 83.5 & 66.3 & 49.4 & 55.9 \\
6 & Podlab & SVM, features & 74.3 & 83.8 & 64.5 & 42.9 & 39.5 \\
% 7 & Politehnica & CNN, transfer   & 71.5 & 81.1 & 61.6 & 38.7 & 38.6 \\
7 & CYUT    & bi-GRU       & 60.3 & 69.8 & 52.4 & 44.3 & 34.4 \\  
\bottomrule
\end{tabular}
\caption{F-scores for Friends (\%)}
\label{tab:friends}

\end{table*}
% Please add the following required packages to your document preamble:
% \usepackage{booktabs}
\begin{table*}[t]
\centering
\begin{tabular}{@{}lllcrrrr@{}}
\toprule
\textbf{Rank} & \textbf{Team} & \textbf{Approach} & \textbf{F1-micro} & \textbf{Neutral} & \textbf{Joy} & \textbf{Sadness} & \textbf{Anger} \\ 
\midrule
1 & IDEA  & BERT, speaker, transfer, context & 88.5 & 92.7 & 77.7 & 60.4 & 47.4 \\
2 & KU & BERT, context, ensemble & 87.6 & 92.2 & 73.9 & 61.8 & 40.0 \\
3 & HSU & BERT & 86.2  & 91.3 & 73.8 & 62.7 & 39.1 \\
4 & Podlab & SVM, features & 85.7 & 91.3 & 67.8 & 52.0 & 22.9 \\
% 5 & Politehnica    & CNN, transfer & 84.2 & 90.0 & 69.0 & 54.5 & 39.0 \\
5 & AlexU & BERT, majority classifier & 84.1 & 90.1 & 75.8 & 52.4 & 30.8 \\
6 & CYUT           & bi-GRU & 71.5 & 81.1 & 58.8 & 28.6 & 10.5 \\
7 & Antenna        & BERT+GRU & 70.2 & 82.2 & 27.9 & 17.9 & 6.7 \\ 
\bottomrule
\end{tabular}
\caption{F-scores for EmotionPush (\%)}
\label{tab:emotionpush}
\end{table*}

A total of eleven teams submitted their evaluations, and are presented in the online leaderboard. Seven of the teams also submitted technical reports, the highlights of which are summarized below. More details are available in the relevant reports.

\paragraph{IDEA} \cite{ex-idea2019} Two different BERT models were developed. For Friends, pre-training was done using a sliding window of two utterances to provide dialogue context. Both Next Sentence Prediction (NSP) phase on the complete unlabeled scripts from all 10 seasons of Friends, which are available for download. In addition, the model learned the emotional disposition of each of six main  six main characters in Friends (Rachel, Monica, Phoebe, Joey, Chandler and Ross) by adding a special token to represent the speaker. 
For EmotionPush, pre-training was performed on Twitter data, as it is similar in nature to chat based dialogues. In both cases, special attention was given to the class imbalance issue by applying ``weighted balanced warming'' on the loss function.
\paragraph{KU}
\cite{ex-ku2019}
BERT is post-trained via Masked Language Model (MLM) and Next Sentence Prediction (NSP) on a corpus consisting of the complete and augmented dialogues of Friends, and the EmotionPush training data.
The resulting token embeddings are max-pooled and fed into a dense network for classification. A $K$-fold cross-validation ensemble with majority voting was used for prediction. 
To deal with the class imbalance problem, weighted cross entropy was used as a training loss function. 

\paragraph{HSU}
\cite{ex-hsu2019}
A pre-trained BERT is fine-tuned using filtered training data which only included the desired labels.
Additional augmented data with joy, sadness, and anger labels are also used. 
BERT is then fed into a standard feed-forward-network with a softmax layer used for classification. 

\paragraph{Podlab}
\cite{ex-adeleide2019}
A support vector machine (SVM) was used for classification. Words are ranked using a per-emotion TF-IDF score. 
Experiments were performed to verify whether the previous utterance would improve classification performance. Input to the Linear SVM was done using one-hot-encoding of top ranking words.

\paragraph{AlexU} \cite{ex-alexandria2019}
The classifier uses a pre-trained BERT model followed by a feed-forward neural network with a softmax output. 
Due to the overwhelming presence of the neutral label, a classifying cascade is employed, 
where the majority classifier is first used to decide whether the utterance should be classified with ``neutral'' or not. 
A second classifier is used to focus on the other emotions (joy, sadness, and anger). 
Dealing with the imbalanced classes is done through the use of a weighted loss function.

\paragraph{Antenna} \cite{ex-antenna2019}
BERT is first used to generate word and sentence embeddings for all utterances. The resulting calculated word embeddings are fed into a Convolutional Neural Network (CNN), and its output is then concatenated with the BERT-generated sentence embeddings. The concatenated vectors are then used to train a bi-directional GRU with a residual connection followed by a fully-connected layer, and finally a softmax layer produces predictions.
Class imbalance is tackled using focal loss \cite{lin2017focal}.

% \paragraph{Politehnica} \cite{ex-politehnica2019}
% Three CNN models independently pre-trained on three different datasets: 
% Friends/EmotionPush, 
% sentiment-labeled data from SemEval-2017 Task 4A \cite{rosenthal2017semeval}, 
% and emotion-labeled data from SemEval-2019 Task 3.

% The input to each of the CNNs are word2vec pre-trained embeddings.
% The encoded representations from each CNN are concatenated and fed into a dropout layer followed by a dense layer and a softmax classifier.

\paragraph{CYUT} \cite{ex-cyut2019} 
A word embedding layer followed by a bi-directional GRU-based RNN. Output from the RNN was fed into a single-node classifier. The augmented dataset was used for training the model, but ``neutral''-labeled utterances were filtered to deal with class imbalance. 

\section{Results}
The submissions and the final results are summarized in Tables \ref{tab:friends} and \ref{tab:emotionpush}. 
Two of the submissions did not follow up with technical papers and thus they do not appear in this summary. 
We note that the top-performing models used BERT, reflecting the recent state-of-the-art performance of this model in many NLP tasks.
For Friends and EmotionPush the top micro-F1 scores were 81.5\% and 88.5\% respectively.

\section{Evaluation \& Discussion}
An evaluation summary of the submissions is available in Tables \ref{tab:friends} and \ref{tab:emotionpush}. We only present the teams that submitted technical reports. A full leaderboard that includes all the teams is available on the challenge website. This section highlights some observations related to the challenge.
Identical utterances can convey different emotions in different contexts. A few of the models incorporated the dialogue context into the model, such as the models proposed by teams \textbf{IDEA} and \textbf{KU}. 

\paragraph{Deep Learning Models.}
Most of the submissions used deep learning models. Five of the models were based on the BERT architecture, with some using pre-trained BERT. Some of the submissions enhanced the model by adding context and speaker related encoding to improve performance. We also received submissions using more traditional networks such as CNN, as well as machine learning classics such as SVM. The results demonstrate that domain knowledge, feature engineering, and careful application of existing methodologies is still paramount for building successful machine learning models.

\paragraph{Unbalanced Labels.} Emotion detection in text often suffers from a data imbalance problem, our datasets included. The teams used two approaches to deal with this issue. Some used a class-balanced loss functions while others under-sampled classes with majority label ``neutral''. Classification performance of underrepresented emotions, especially sadness and anger, is low compared to the others. This is still a challenge, especially as some real-world applications are dependent on detection of specific emotions such as anger and sadness.

\paragraph{Emotional Model and Annotation Challenges.} 
% Please add the following required packages to your document preamble:
% \usepackage{booktabs}
% Please add the following required packages to your document preamble:
% \usepackage{booktabs}
\begin{table}[hpt!]
\centering
\begin{tabular}{@{}lccc@{}}
\toprule
 & \textbf{Friends} & \textbf{EmotionPush} & \textbf{Both}\\ 
\midrule
Joy & 0.401 & 0.454 & 0.429 \\
Neutral & 0.350 & 0.349 & 0.373 \\
Surprise & 0.330 & 0.266 & 0.322 \\
Sadness & 0.291 & 0.324 & 0.307 \\
Anger & 0.319 & 0.212 & 0.306 \\
Disgust & 0.184 & 0.160 & 0.179 \\
Fear & 0.184 & 0.132 & 0.176 \\
\bottomrule
\end{tabular}
\caption{Per-emotion Reliability of Agreement ($\kappa$) }
\label{tab:irr_emotion}
\end{table}
The discrete 6-emotion model and similar models are often used in emotion detection tasks. However, such 1-out-of-n models are limited in a few ways: first, expressed emotions are often not discrete but mixed (for example, surprise and joy or surprise and anger are often manifested in the same utterance). 
This leads to more inter-annotator disagreement, as annotators can only select one emotion.
Second, there are additional emotional states that are not covered by the basic six emotions but are often conveyed in speech and physical expressions, such as desire, embarrassment, relief, and sympathy. 
This is reflected in feedback we received from one of the AMT workers: ``\textit{I am doing my best on your HITs.  However,  the emotions given (7 of them) are a lot of times not the emotion I'm reading (such as questioning, happy, excited, etc).  Your emotions do not fit them all...}''.

To further investigate, we calculated the per-emotion $\kappa$-statistic for our datasets in Table \ref{tab:irr_emotion}. We see that for some emotions, such as disgust and fear (and anger for EmotionPush), the $\kappa$-statistic is poor, indicating ambiguity in annotation and thus an opportunity for future improvement. We also note that there is an interplay between the emotion label distribution, per-emotion classification performance, and their corresponding $\kappa$ scores, which calls for further investigation.

\paragraph{Data Sources.} One of the main requirements of successful training of deep learning models is the availability of high-quality labeled data. Using AMT to label data has proved to be useful. However, current data is limited in quantity. In addition, more work needs to be done in order to measure, evaluate, and guarantee annotation quality. In addition, the Friends data is based on an American TV series which emphasizes certain emotions, and it remains to be seen how to transfer learning of emotions to other domains.

\section{Acknowledgment}
This research is partially supported by Ministry of Science and Technology, Taiwan, under Grant no. MOST108-2634-F-001-004- and MOST107-2218-E-002-009-.
\bibliography{library} 
\bibliographystyle{named}
\end{document}